\documentclass[10pt, conference, compsocconf]{IEEEtran}
\usepackage{times}
\usepackage{epsfig}
\usepackage{graphicx}
\usepackage{amsmath}
\usepackage{amssymb}
\usepackage{times}
\usepackage{colortbl}
\usepackage{subfigure}
\usepackage{slashbox}
\usepackage{capt-of}

\begin{document}
\title{Fusing Face and Periocular biometrics using Canonical correlation analysis}
\author{N. S. Lakshmiprabha\\
}

\maketitle
\thispagestyle{empty}

\begin{abstract}
This paper presents a novel face and periocular biometric fusion at feature level using canonical correlation analysis. Face recognition itself has limitations such as illumination, pose, expression, occlusion etc. Also, periocular biometrics has spectacles, head angle, hair and expression as its limitations. Unimodal biometrics cannot surmount all these limitations. The recognition accuracy can be increased by fusing dual information (face and periocular) from a single source (face image) using canonical correlation analysis (CCA). This work also proposes a new wavelet decomposed local binary pattern (WD-LBP) feature extractor which provides sufficient features for fusion. A detailed analysis on face and periocular biometrics shows that WD-LBP features are more accurate and faster than local binary pattern (LBP) and gabor wavelet. The experimental results using Muct face database reveals that the proposed multimodal biometrics performs better than the unimodal biometrics. 
\end{abstract}


\section{Introduction}
\label{intro}
Face recognition is an active research for past four decades. Still there are unaddressed challenges put forth infront of researchers. Face recognition has its wide application in security, surveillance and authentication fields. Unlike iris and fingerprint recognition, face recognition does not require user cooperation. Faster development in technologies is also resulted in hacking, filching and other fraudulent activities. Plastic surgery and spoofing (photo attack) are the two new hurdles, allowing fugitives to roam freely without any fear about the face recognition system. Facial aging is another issue which affects accuracy as well as performance of face recognition. Some of these challenges can be addressed by using dual information from different biometrics system.    

Periocular (a region surrounding eyes), is considered as most discriminative in nature \cite{periocular1}. This biometrics faces problem in presence of head angle or pose variations, spectacles, hair and facial expression \cite{mypaper}. Thus the periocular biometrics cannot serve as an individual biometrics \cite{perio1}. Fusing periocular region data with other biometrics may help in increasing total recognition rate. Obtaining dual information from a single source reduces the total system cost and data acquisition time to some extent. Fusing face and periocular biometrics obtained from same source is more attractive.

Fusion can occur at, image level, feature level, match score level and decision level \cite{multi1}. In \cite{mypaper}, face and periocular is fused at decision level. This paper fuses face and periocular biometrics at feature level using Canonical Correlation Analysis (CCA) \cite{CCA1}. CCA tries to find the maximum correlation between two feature set which helps in increasing the accuracy and precision. Feature extraction step plays an important role in CCA. Local Binary Pattern (LBP) is a well known feature extractor for both periocular and face images \cite{mypaper,lpbref1}. LBP on all wavelet sub-bands are used to resolve writer identification problem \cite{wdlbp}. LBP provides rich texture features by considering more number of blocks in a given image. Increase in number of blocks also increases computational complexity and processing time. This is overcome by performing LBP only on approximate part of the wavelet decomposed image \cite{WDref2}. Further to reduce the size of feature, Principle Component Analysis (PCA) is used. Slight variation on lighting condition causes huge variation in PCA results. Calculating principle component on extracted feature instead on direct intensity value helps in increasing the results. Thus this paper proposes, wavelet decomposed LBP (WD-LBP) based multimodal biometrics from face and periocular images fusion using CCA. Muct face database \cite{muctdb} is used for evaluating the performance of proposed methods, since it has multi-ethnic face images with other variations such as illumination, pose, expression and occlusion. 
 
This paper begins with the detailed description of proposed method (see section \ref{proposed}), followed by feature extraction using wavelet decomposed LBP in section \ref{Featureextract}. Face and periocular feature fusion using CCA is explained in section \ref{CCA}. Section \ref{classi}, elaborates on classification step. The experiments performed using Muct face database is analyzed in section \ref{results}. Finally, conclusion is presented in section \ref{conclude}.

\section{Proposed Method}
\label{proposed}
\begin{figure}[htp]
\centering
\includegraphics[width=8.5cm]{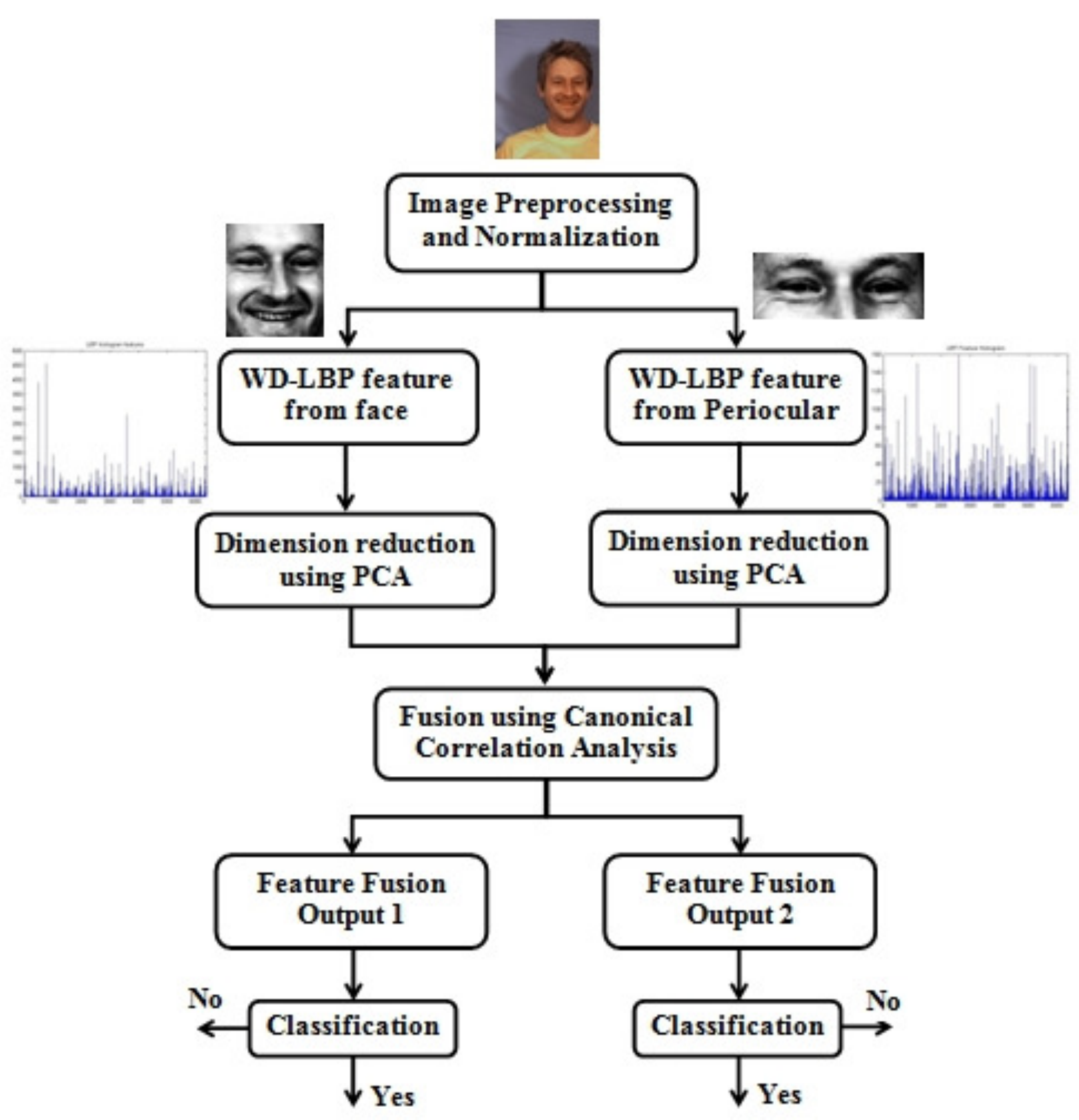}
\caption{Block diagram of the proposed multimodal biometrics using Canonical Correlation Analysis (CCA)}
\label{fig1}
\end{figure}

Local binary pattern (LBP) extracts local information by generating a pattern using binary values in a particular region. In order to increase the strength of feature set, an image is divided into number of blocks. LBP performs better with increased number of blocks on larger size image. This also increases the computational complexity as well as processing time. It is always required to fetch as much information possible from a smaller size image. However, normalizing an image to much smaller size causes huge information loss. This paper illustrates two ways to overcome above said issues, 1) LBP is performed on the approximate part of wavelet decomposed image - aids in improving the accuracy with less number of blocks, 2) On the LBP extracted feature, PCA is performed for feature dimension reduction - which reduces computational complexity and processing time during classification. PCA also helps in making these local features invariant to noise. The recognition rate will also increase because the problem of PCA under illumination variation is eliminated by these features. Further to increase the accuracy, face and periocular biometrics is fused using canonical correlation analysis (CCA). Periocular is one of other discriminative features in face. Thus by fusing two biometrics data (i.e. face and periocular) obtained from the same source (high resolution face image) by finding maximum correlation may increase the total performance of the system. Block diagram of the proposed method is shown in figure \ref{fig1}. Periocular biometric can be performed in 3 different ways such as overlapping, Non-overlapping, Strip. These three regions can be obtained by taking four eye corner points \cite{mypaper}. This paper considers strip type without placing mask in the eye region assuming eyes are open in all the images. Two fusion outputs produced from CCA is analyzed separately for its performances.    

\section{Feature Extraction using Wavelet Decomposed LBP (WD-LBP)}
\label{Featureextract}
\begin{figure}[htp]
\centering
\includegraphics[width=8.5cm]{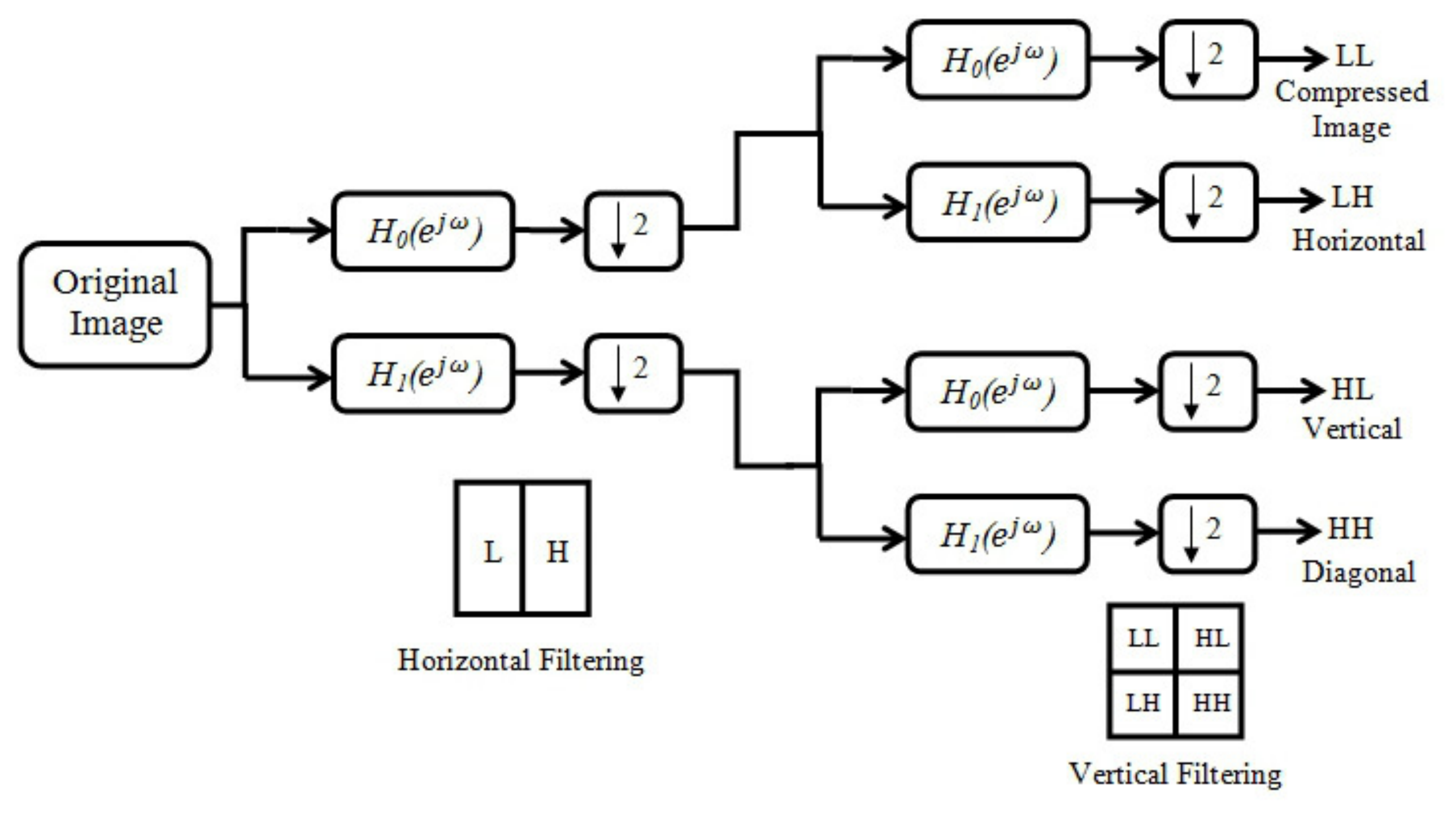}
\caption{The block diagram depicting the operation of wavelet decomposition}
\label{wave}
\end{figure}

Wavelet decomposition (WD) is widely used for image compression. WD down-samples input image by a factor of two in each level decomposition \cite{WDref2} with less information loss. Filter banks are elementary building blocks of wavelets which produces four sub-bands (A1 - approximate, H1 - horizontal, V1 - vertical and D1 - diagonal) as its first level output. An analysis filter bank consist of a low pass filter $H_0(e^{j\omega})$, a high pass filter $H_1(e^{j\omega})$ and down-samplers \cite{WDref2}. These filter banks are cascaded to form wavelet decomposition (WD). Approximate part from first level decomposition (A1) is further decomposed into second level and this is carried out to several levels. This paper uses Daubechies wavelet 8 with two level of decomposition for all experiments. The approximate part from second level decomposition (A2) is used in LBP feature extraction. The steps involved in wavelet decomposition is shown in figure \ref{wave}.

Local Binary Pattern (LBP) provides rich texture features which represents an object in meaningful way. LBP features are gray scale and rotation invariant texture operator \cite{lpbref1}. LBP features can be extracted faster than Gabor wavelets and its performance is almost similar.

\begin{figure}[htp]
\centering
\includegraphics[width=8.5cm]{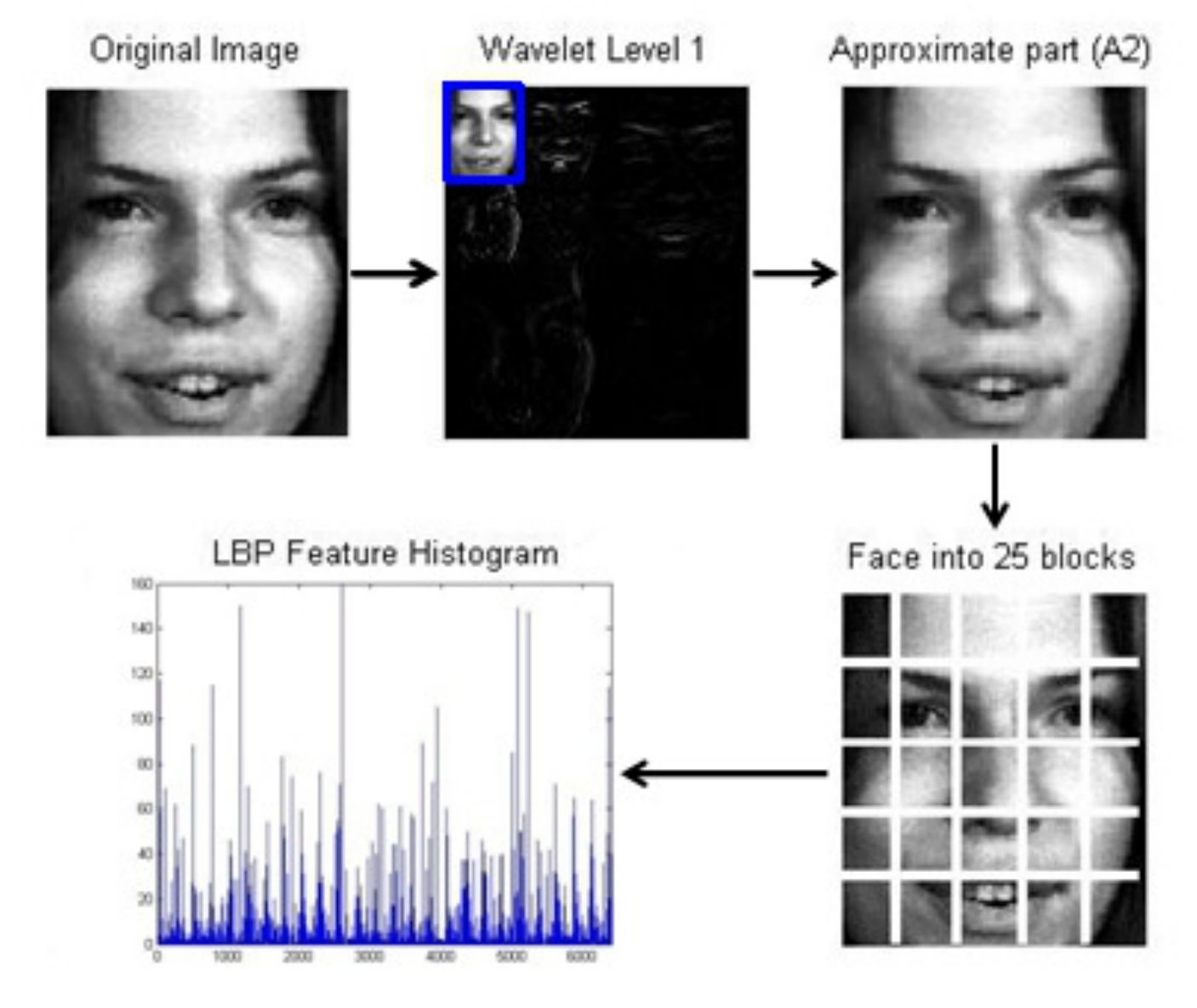}
\caption{Feature extraction using wavelet decomposed LBP (WD-LBP)}
\label{fig2}
\end{figure}

Consider a 3x3 window with center pixel $(x_c, y_c)$ intensity value be $g_c$ and local texture as $T = t(g_i)$ where $g_i(i = 0, 1, 2, 3, 4, 5, 6, 7)$ corresponds to the grey values of the 8 surrounding pixels. These surrounding pixels are thresholded with the center value $g_c$ as $t(s(g_0 - g_c), \dots , s(g_7 - g_c))$ and the function s(x) is defined as,
\begin{equation}
s(x) = \left\{
\begin{array}{lr}
1 & , x > 0\\
0 & , x \le 0
\end{array}
\right.
\end{equation} 
The LBP pattern at the center pixel $g_c$ can be obtained using equation (\ref{lbp2}). Figure \ref{fig2} shows various steps involved in wavelet decomposed LBP feature extractor. In order to increase the feature strength and to get more details, the face images are divided into number of blocks. Face image with five number of divisions along row and column wise which results in total 25 blocks is shown in figure \ref{fig2}.
\begin{equation}
LBP(x_c, y_c) = \sum_{i=0}^{7} s(g_i - g_c) 2^i
\label{lbp2}
\end{equation}

Increasing the number of blocks in image, the feature histogram also increases. Five division (25 blocks) along row and column wise results in feature vector of 6400 x 1 pixels whereas six division (36 blocks) gives 9216 x 1 pixels for an image. This huge increase in feature vector size can be reduced using Principal Component Analysis (PCA) \cite{pca}. The feature vector can be reduced to the size less than or equal to number of images in the training set.

Let $I=[i_1, i_2,\dots,i_M]$ be the WD-LBP features for M training set images. Each feature vector in I can be projected into the eigenface space $\omega$ using equations (\ref{projection}). 
\begin{equation}
\omega = u . \phi = u . (i_k - \psi) 
\label{projection}
\end{equation}
where $u$ is the eigenvectors of the covariance matric $C = \frac{1}{M} \sum_{n=1}^M\phi_n\phi_n^T = AA^T$, $\psi$ is the mean face $\psi= \frac{1}{M}\sum_{n=1}^Mi_n$, k=1,2,\dots,M and $\phi$ is the mean substracted face images. 

Weight Matrix $\Omega = [\omega_1, \omega_2, \ldots , \omega_M']^T$ is the representation of WD-LBP training set features in eigenface space.

\section{Fusion using Canonical Correlation Analysis}
\label{CCA}
The features obtained from face and periocular regions are fused using canonical correlation analysis (CCA). The goal here is to get the maximum variation out of the two feature vectors \cite{CCA1,CCA}. Let there are N number of images in the training set, feature from face be $X = [x_1, x_2, \dots x_N]$ and periocular image be $Y = [y_1, y_2, \dots y_N]$ with $n_1$ and $n_2$ as its dimension, where $n_1, n_2 \le N-1$. $\overline{x}_i$ and $\overline{y}_i$ be the mean subtracted (i.e. zero means) feature vector. The covariance matrix, $$C_{xx} = \frac{1}{N}\sum_{i=1}^N \overline{x}_i\overline{x}_i^T = \frac{1}{N}\overline{X}*\overline{X}^T$$ similarly $C_{yy}$ and $C_{xy}$ can be calculated.    
\begin{eqnarray}
\label{cca3}
C_{xx}^{-1}C_{xy}C_{yy}^{-1}C_{xy}^Ta = \rho^2a\\
\label{cca4}
C_{yy}^{-1}C_{xy}^TC_{xx}^{-1}C_{xy}b = \rho^2b
\end{eqnarray}
where, a and b are the eigenvectors and $\rho$ being the eigenvalues or correlation coefficients or canonical correlations. Let A and B be the canonical basis vector of feature set X and Y.

Given $x_i$ and $y_i$ be the features obtained from face and periocular data respectively. These independent features can be fused by projecting it to the canonical basis vectors using equation (\ref{cca5}) or (\ref{cca6}). $z_{i1}$ and $z_{i2}$ is referred as feature fusion output 1 (FFO 1) and feature fusion output 2 (FFO 2) respectively.
\begin{eqnarray}
\label{cca5}
z_{i1} = \left[
\begin{array}{c}
A^T\overline{x}_i\\
B^T\overline{y}_i
\end{array}
\right]\\
\label{cca6}
z_{i2} = \left(
A^T\overline{x}_i + B^T\overline{y}_i
\right)
\end{eqnarray}

\section{Classification}
\label{classi}
In order to avoid the time consumption in training and testing, euclidean distance classifier is used. $z_{kj}$ of the test images is compared with each of the $z_{ij}$ in the training set, where i = 1, 2, \dots N and j = 1 or 2 (i.e. feature fusion output 1 or 2) using euclidian distance, $\varepsilon_i$.
\begin{equation}
\varepsilon_i^2 = ||z_{j} - z_{ij}||^2
\end{equation}
If minimum of $\varepsilon_i$ is below threshold value $\theta$, then ith image in the training set is declared as the matching one. Threshold value can be calculated using, $\Theta= \frac{1}{2} max(||z_{pj} - z_{qj}||)$, where p and q are images from same class.
\section{Experimental Results and Discussion}
\label{results}
Experiments are carried out using 500 face images (10 images from each individual) from Muct database \cite{muctdb}. Among 500 images, 250 images (5 images per person) are used for training and remaining for validating the algorithm. The face image is normalized to a size of 150x130 pixels and the size of periocular (strip) region is 50x130 pixels. The size of face and periocular image after two level of WD is 48x43 and 23x43 pixels respectively. The recognition rate obtained from face images using LBP and LBP-PCA is 72(\%) and 81.6(\%) with 0.219 and 0.007 seconds as classification time for an image. PCA helps in increasing the recognition rate with less classification time. Thus, PCA is carried out as feature dimension reduction step in all these methods. 

Table \ref{table1} compares the result obtained from validation set images using Gabor, LBP and WD-LBP feature extractors. Gabor wavelet is performed with five scales and eight orientations. The recognition rates obtained from periocular region is better than face which clearly shows the discriminative nature of periocular region. In case of face image, the time taken for training and testing all the images (i.e. 500) with 9 number of divisions using LBP features is 107 seconds and WD-LBP features is 91 seconds. The experiments are conducted with 4 GB RAM and 2.40 GHz speed personal computer using MATLAB 7.0 software. The results from LBP and WD-LBP are obtained with 12 and 9 number of divisions (along row and column wise) respectively. WD-LBP feature gives better results than LBP even with less number of divisions. This helps in attaining good recognition rate with less processing time. 
\begin{table}[htp]
\centering
\begin{tabular}{|c|c|c|c|}
\hline
\rowcolor[gray]{.8}
Methods & Gabor & LBP & WD-LBP\\
\hline
Face & 80.8 & 81.6 & 84\\
\hline
Periocular & 78 & 82.8 & 85.6\\
\hline
\end{tabular}
\caption{Comparison of results obtained from Gabor, LBP and WD-LBP Feature extractors.}
\label{table1}
\end{table}
\begin{figure}[htp]
\centering
\includegraphics[width=8.5cm]{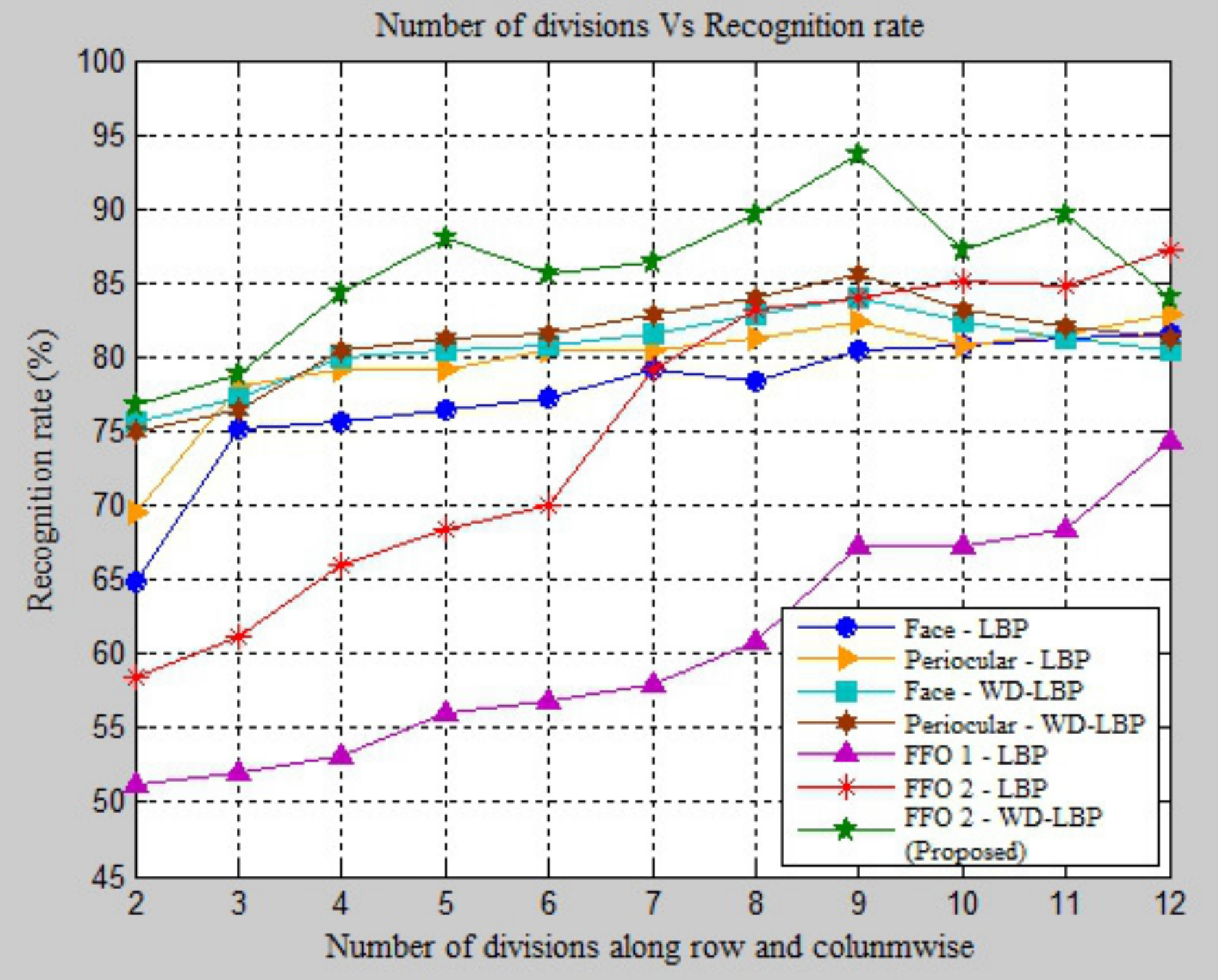}
\caption{Number of divisions along row and column wise Vs Recognition rate using LBP, WD-LBP and fusion methods.}
\label{fig3}
\end{figure}

Figure \ref{fig3} shows the recognition rates obtained with different number of divisions along row and column wise using LBP and WD-LBP features. It is observed from the plot that feature fusion output 1 (FFO 1) performance is unacceptable whereas feature fusion output 2 (FFO 2) performs better using LBP features. Thus, the proposed WD-LBP features from face and periocular is fused only with FFO 2 method. Results from multimodal fusion with FFO 2 method is also given in table \ref{table2} for better understanding. 
\begin{table}[htp]
\centering
\begin{tabular}{|c|c|c|c|c|c|c|c|c|}
\hline
\rowcolor[gray]{.8}
No.of Div. & 7 & 8 & 9 & 10 & 11 & 12\\
\hline
LBP & 79.2 & 83.2 & 84 & 85.2 & 84.8 & \bf 87.2\\
\hline
WD-LBP & 86.4 & 89.6 & \bf 93.6 & 87.2 & 89.6 & 84\\
\hline
\end{tabular}
\caption{Results from multimodal (face and periocular) fusion using CCA with different number of divisions along row and column wise with LBP and WD-LBP features.}
\label{table2}
\end{table}

WD-LBP based multimodal fusion using CCA performs the best when compared to all other discussed methods. There is a peak in all WD-LBP method after which the recognition rate starts decreasing. When the number of divisions increases, the size of each block becomes too small. For example, after two levels of WD and with 12 number of division, the size of block is 4x3 pixels for face and 1x3 pixels for periocular. Performing LBP at this smaller size is meaningless. This forms the reason behind decrease in the recognition rate after a peak. 
\section{Conclusion}
\label{conclude}
The experimental results obtained using muct database shows that the periocular biometrics performs better than face biometrics. PCA as feature dimension reducer steadily increases the recognition rate with less classification time. The performance of WD-LBP is better and faster than LBP and Gabor features. Further increase in recognition rate is observed from feature fusion output 2 (FFO 2) method than feature fusion output 1 (FFO 1) method. The proposed WD-LBP feature based face and periocular biometrics fusion using CCA gives the best results compared to other fusion and individual biometrics methods.
{
\bibliographystyle{IEEEtran}
\bibliography{Fusion1}
}
\end{document}